\documentclass[12pt]{article}
\usepackage{amsmath}
\usepackage{latexsym}
\usepackage{amsfonts}
\usepackage[normalem]{ulem}
\usepackage{array}
\usepackage{amssymb}
\usepackage{graphicx}
\usepackage{subcaption}
\usepackage[backend=biber,style=authoryear, url=false, doi=false]{biblatex}
\usepackage[bottom]{footmisc}
\usepackage[svgnames,table]{xcolor}
\usepackage{hyperref}
\usepackage[toc,page]{appendix}
\usepackage[paperheight=11.0in,paperwidth=8.5in,left=1.0in,right=1.0in,top=1.0in,bottom=1.0in,headheight=1in]{geometry}

\title{Considerations for the Interpretation of Bias Measures of Word Embeddings}
\date{}
\author{
  Inom Mirzaev$^*$\\
  \texttt{inom.mirzaev@workday.com}
  \and
  Anthony Schulte\thanks{The two first authors contributed equally.}\\
  \texttt{tony.schulte@workday.com}
  \and
  Michael Conover\\
  \texttt{michael.conover@workday.com}
 \and
  Sam Shah\\
  \texttt{shahsam@umich.edu}
}

\begin{document}
\maketitle
\begin{abstract}
    Word embedding spaces are powerful tools for capturing latent semantic relationships between terms in corpora  \autocite{mikolovEfficientEstimationWord2013a,penningtonGloveGlobalVectors2014}, and have become widely popular for building state-of-the-art natural language processing algorithms. However, recent studies have shown that societal biases (e.g., gender, race, age, etc.) present in text corpora may be incorporated into the word embedding spaces learned from them as well. Thus, there is an ethical concern that human-like biases contained in the corpora and their derived embedding spaces might be propagated, or even amplified with the usage of the biased embedding spaces in downstream applications. In an attempt to quantify these biases so that they may be better understood and studied, several bias metrics have been proposed. We explore the statistical properties of these proposed measures in the context of their cited applications as well as their supposed utilities. We find that there are significant caveats to the simple interpretation of these metrics as proposed, and that some applications of these metrics in well-cited works may be erroneous. Specifically, we find that the bias metric proposed by \autocite{bolukbasiManComputerProgrammer2016} is highly sensitive to embedding hyper-parameter selection, and that in many cases, the variance due to the selection of  some hyper-parameters, notably the embedding space dimensionality, is greater than the variance in the metric due to corpus selection, while in fewer cases, even the relative rankings of the bias measured in the embedding spaces of various corpora varies with hyper-parameter selection. In light of these observations, it may be the case that bias estimates should not be thought to directly reflect the properties of the underlying corpus, but rather the properties of the specific embedding spaces in question, particularly in the context of hyper-parameter selections used to generate them. Hence, bias metrics of spaces generated with differing hyper-parameters should be compared only with explicit consideration of the embedding-learning algorithms' particular configurations. 
    \par While it may be possible to use embedding spaces generated with a controlled hyper-parameter configuration to rank corpora in terms of the quantity of bias contained, the numerical value of this bias metric has poor stability across model configurations and a somewhat unclear interpretation in the context of the hyper-parameter sensitivity, and should not be presented as a canonical measure of the corpus in question without reporting the setup of the embedding-learning algorithm used. Moreover, due to the potentially imprecise means by which the term sets used to identify the bias axes, $G_1$, $G_2$ and $W$, are found, we find it difficult to defend the canonicity of these term sets, and offer that it may be helpful to include statistics related to the distribution of the bias metric under variously sampled metric-inducing term sets to defend against critiques related to the hyper-parametric selection of a particular set of terms in the formulation of the particular bias metric in question.
\end{abstract}

\section*{Introduction}
\addcontentsline{toc}{section}{Introduction}
Word embeddings are widely used for their ability to capture the semantic meanings of terms within a corpus. They are widely praised as useful tools for generating features for use in natural language processing systems, and recently, some researchers are attempting to study the structures of these embedding spaces to learn about the fundamental linguistic properties, or even the semantic content of a corpus. While many of the properties of embedding spaces are useful and socially benign, a subset of these properties can reveal latent relationships between terms which could be socially problematic, and may be of interest to researchers for study, or possibly lead to risks of propagating the implicit biases of a corpus if these socially problematic term relationships are used in downstream machine learning applications which would ideally be free of such implicit biases. Qualitatively, when inspecting term-analogical relationships as is done in \autocite{gargWordEmbeddingsQuantify2018}, it is difficult to deny the existence of these biases, but the task of quantitatively capturing them in a canonical measure has been a topic of recent study. Initial attempts have been made to develop metrics which seek to describe the geometric properties of the embedding space with respect to various axes of interest which are empirically determined to correspond to our intuitions of the hypothetical biases under study to quantify the degrees to which various biases exist within the embedding space, and presumably, the underlying text corpus \autocite{bolukbasiManComputerProgrammer2016, caliskanSemanticsDerivedAutomatically2017, gargWordEmbeddingsQuantify2018}. For instance, using such a bias measure, \autocite{bolukbasiManComputerProgrammer2016} concluded that ``word embeddings trained on Google News articles exhibit female/male gender stereotypes to a disturbing extent''.  To the end of quantifying these biases, \autocite{caliskanSemanticsDerivedAutomatically2017} proposed and used a bias measure to report that ``text corpora contain recoverable and accurate imprints of our historic biases''. The genesis of these biases via the underlying text corpus is explored in depth in \autocite{brunetUnderstandingOriginsBias2018}. Although this family of bias metrics is fairly new, initial attempts have been made to refine and explore their validity and robustness as explored in the metric significance research in \autocite{caliskanSemanticsDerivedAutomatically2017}, which seeks to determine whether a bias score is significant, given the possibility of having selected various sets of metric-inducing terms.

\section*{Properties of the bias measurement technique}
\addcontentsline{toc}{section}{Properties of the bias measurement technique}

In this section, we evaluate the stability of the bias measure developed in \autocite{bolukbasiManComputerProgrammer2016} which is claimed to measure societal biases in word embeddings. For the sake of completeness, we repeat the definition of \autocite{bolukbasiManComputerProgrammer2016}'s bias metric:\newline\newline Given two groups of words (e.g., gender words)
	\begin{equation*}
	    G_1 = \{\Vec{x_i}\}_{i=1}^K \text{\space\space and \space} G_2 = \{\Vec{y_i}\}_{i=1}^K\ ,
	\end{equation*}
	first groups' subspace direction $\Vec{g}$ is identified as the first principal component of the vectors
	\begin{equation*}
	    \{\Vec{x_i}-\Vec{y_i}\}_{i=1}^K \cup \{\Vec{y_i}-\Vec{x_i}\}_{i=1}^K\ .
	\end{equation*}
	Consequently, given a set of words $W=\{\Vec{w_i}\}$, \textit{bias} for each word is defined as
	\begin{equation*}
        cosine\_similarity \left(\Vec{w_i},\Vec{g}\right)
	\end{equation*}
	and \textit{direct bias} for a given set of words is calculated as
	\begin{equation*}
	    \frac{1}{\left|W\right|}\sum\left| cosine\_similarity\left(\Vec{w_i},\Vec{g}\right)\right|\ .
	\end{equation*}

In essence, this bias metric measures how closely a given word embedding aligns with a gender axis defined by a first principal component of vectors \textit{man-woman, he-she, him-her, etc.} And, direct bias is essentially average magnitude of bias present in a list of neutral words.

\begin{figure}
    \centering
    \includegraphics[width=0.8\textwidth]{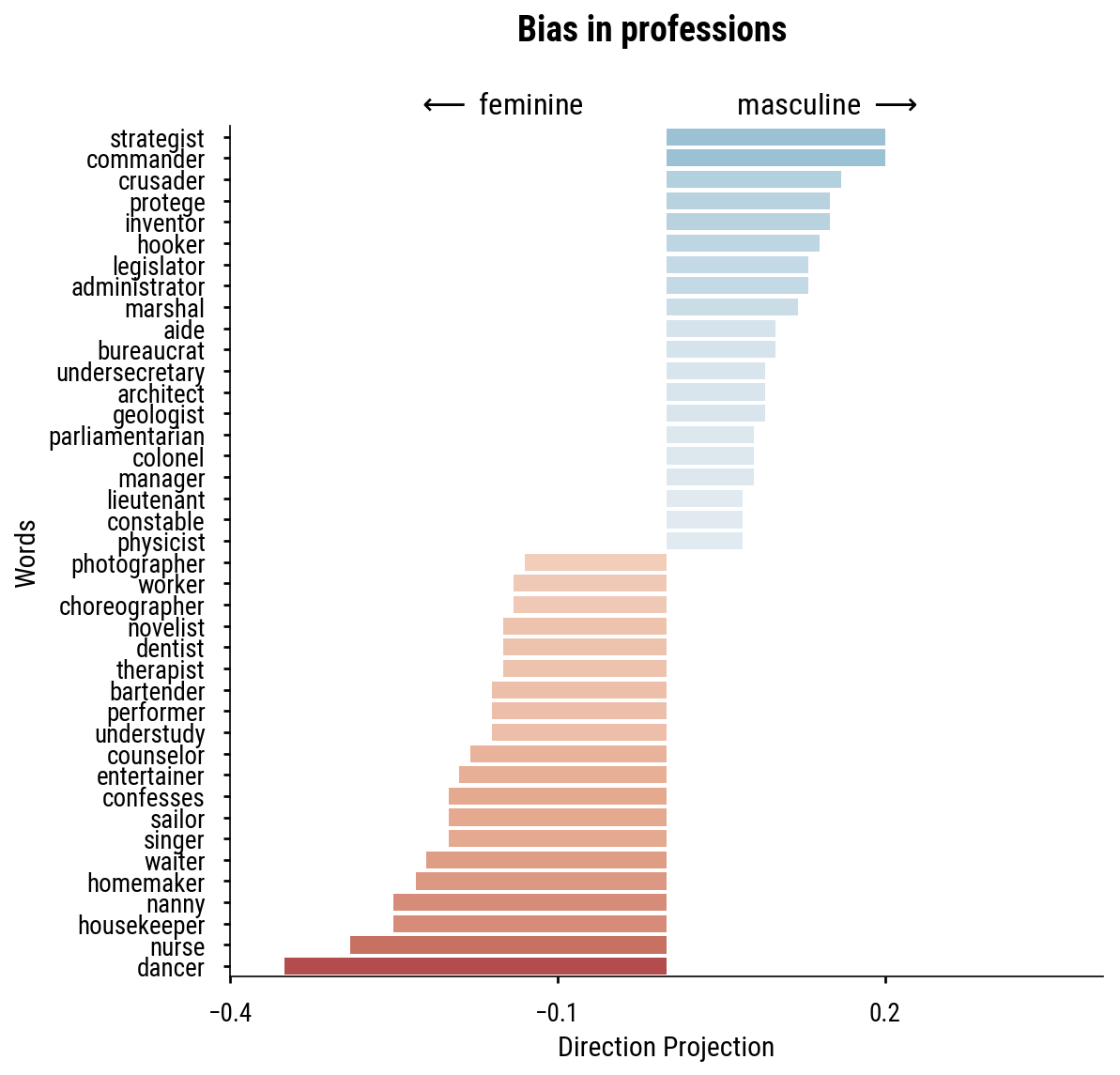}
    \caption{Bias scores for a list of profession names using \autocite{bolukbasiManComputerProgrammer2016} bias measure.}
    \label{fig:sample_biases}
\end{figure}

\begin{figure}
    \centering
    \includegraphics[width=0.8\textwidth]{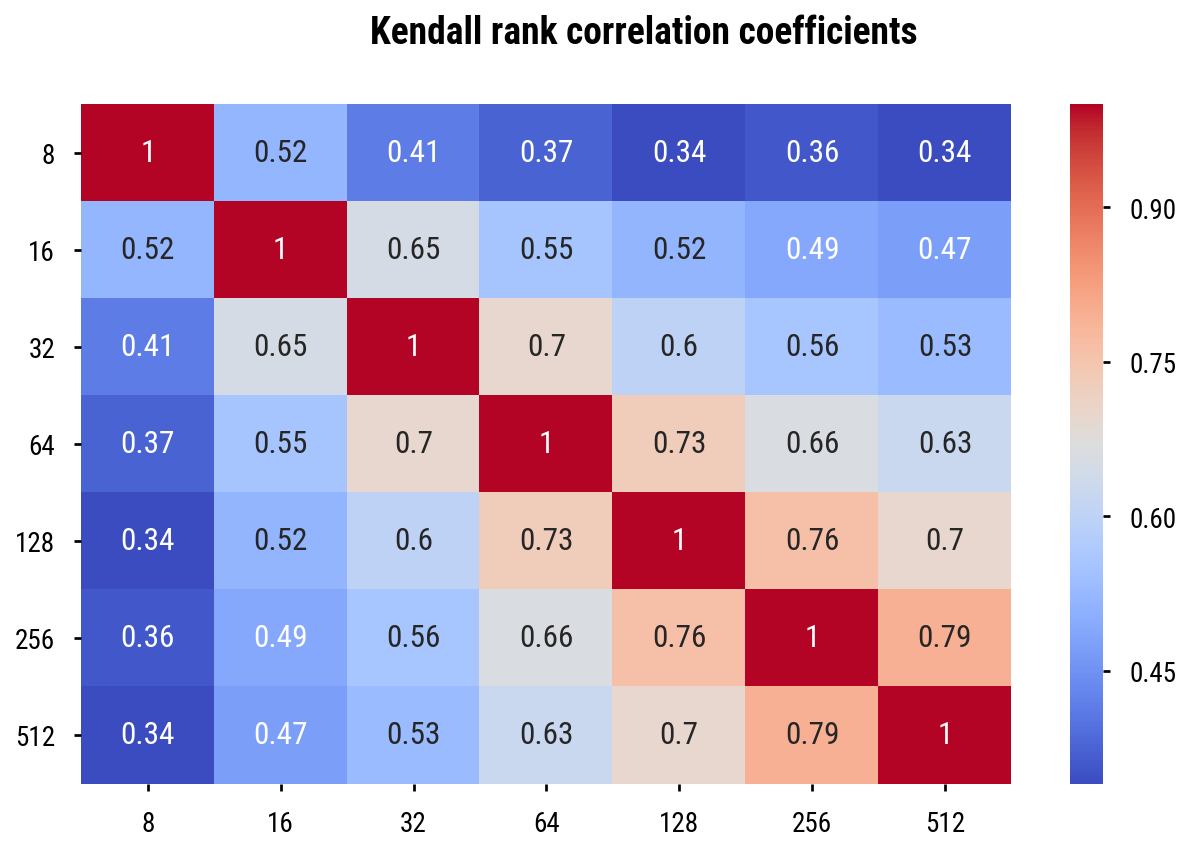}
    \caption{Kendall tau rank correlation coefficients for pairs of word embeddings trained using Skip-gram algorithm.}
    \label{fig:kendall_tau}
\end{figure}

Figure~\ref{fig:sample_biases} illustrates the results of \autocite{bolukbasiManComputerProgrammer2016} bias detection algorithm for a list of profession names with word embedding vectors of dimension 256 trained using the Skip-gram algorithm \autocite{mikolovEfficientEstimationWord2013a} on a sample of 23k Wikipedia articles with 50k term vocabulary. As one can observe, this bias measure identifies some profession names such as commander and nurse, which historically were predominantly male and female jobs, respectively. Moreover, we trained word embeddings with differing dimension on the same sampled Wikipedia corpus using Skip-gram algorithm, and calculated Kendall tau rank correlation coefficients for the bias metrics corresponding to terms in the above list of professions. As illustrated in Figure~\ref{fig:kendall_tau}, we found that though rankings for low dimensional embeddings were unstable, for larger dimensions ($\ge 128$) rankings of the biases of the profession terms achieve superior Kendall tau scores.

\begin{figure}

        \centering
    \begin{subfigure}[t]{0.45\textwidth}
		\includegraphics[width=\textwidth]{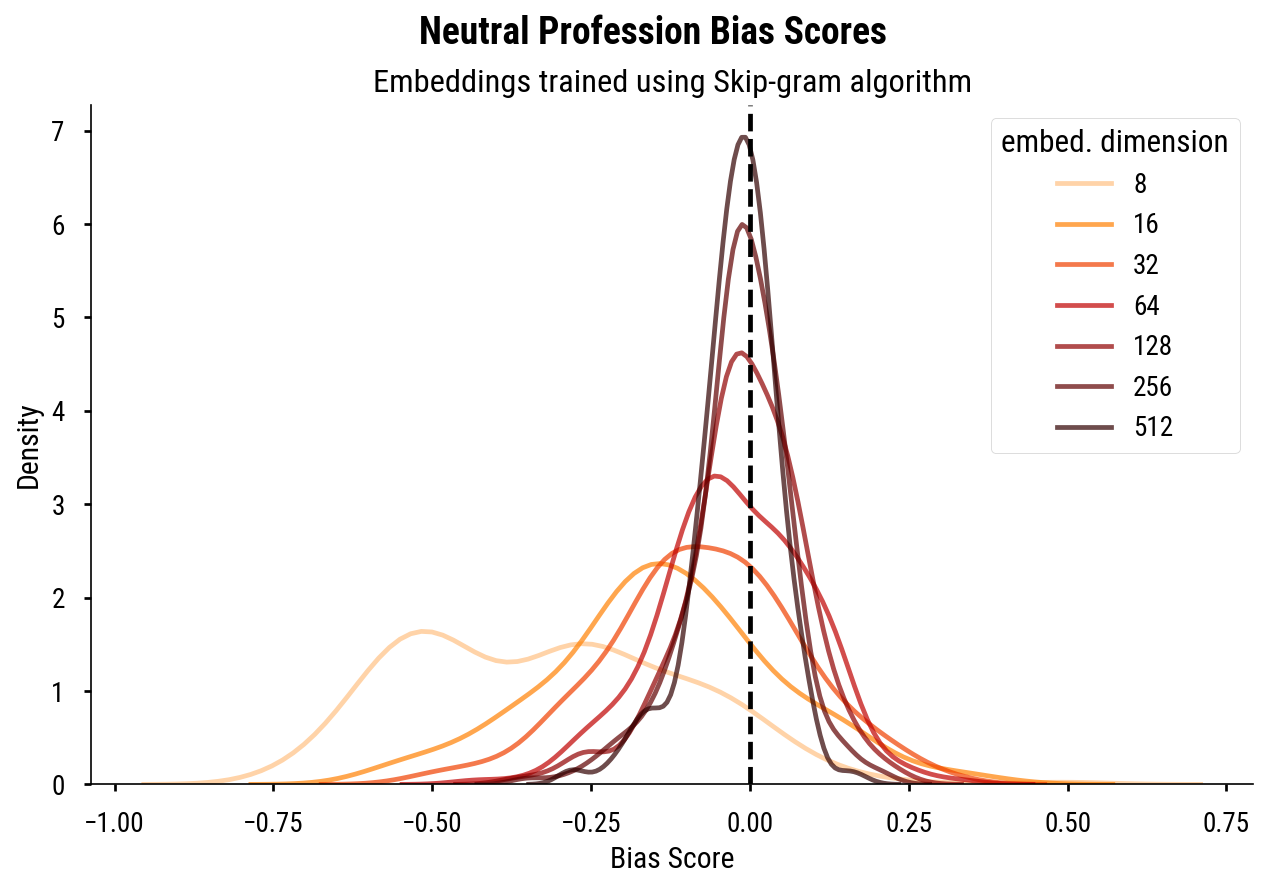}
        \caption{}
        \label{fig:bias_density}
    \end{subfigure}
    ~
    \begin{subfigure}[t]{0.45\textwidth}
		\includegraphics[width=\textwidth]{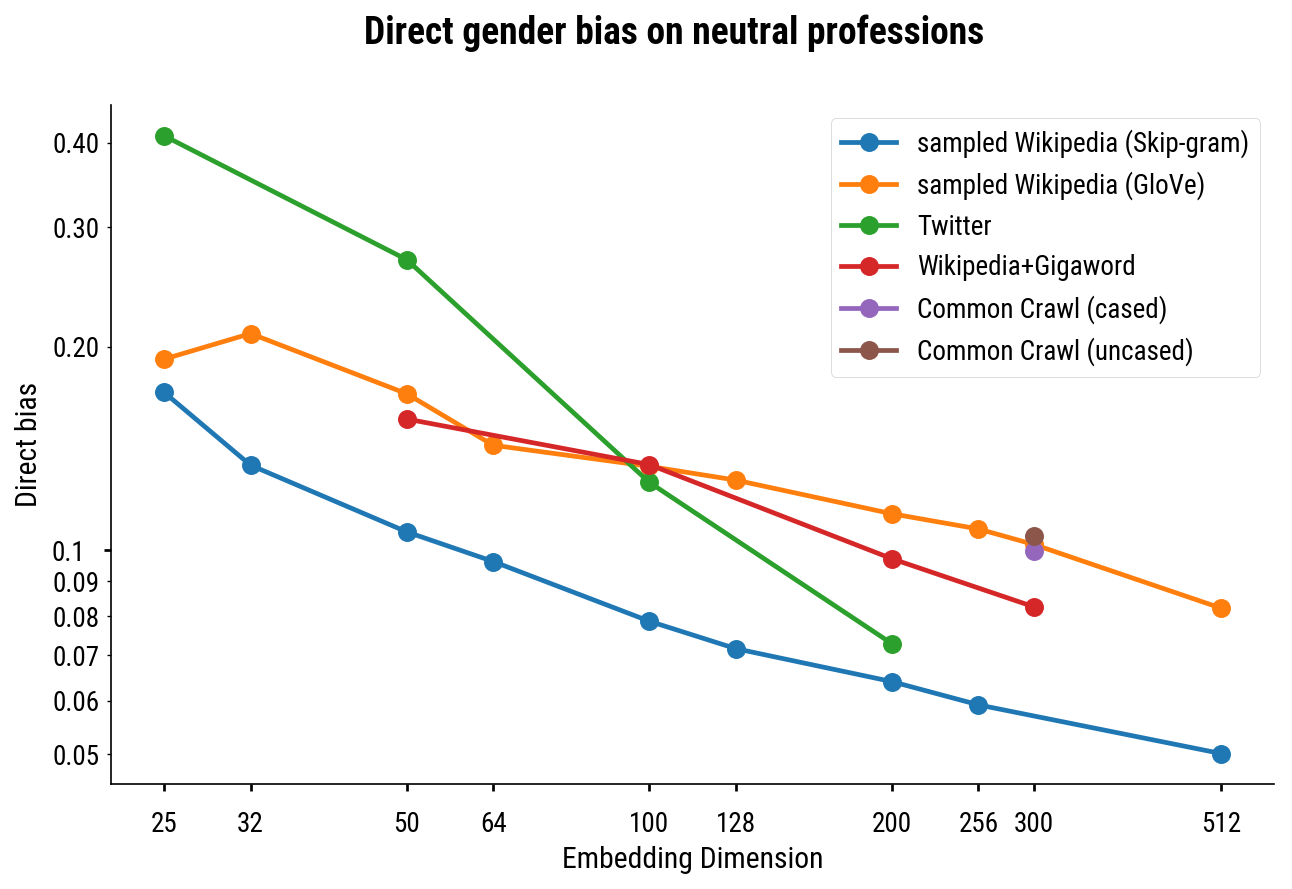}
        \caption{}
        \label{fig:bias_decay}
    \end{subfigure}

    \caption{Bias score decay with increasing dimension \textbf{(a)} Bias score densities for word embeddings trained using Skip-gram \textbf{(b)} Direct bias measures for publicly available embeddings trained with GloVe$^1$ and our embeddings trained on sampled Wikipedia data. Note that the ranking of the twitter corpus changes with embedding dimension.}
\end{figure}

Although this bias measure nearly preserves term-bias-rankings, particularly for highly dimensional embedding configurations, we found that bias measure decays, approaching zero as the space dimensionality increases. In Figure~\ref{fig:bias_density}, we have depicted bias score densities for word embeddings trained for a range of dimensions using Skip-gram algorithm on the same text corpus. One can observe that bias score densities change substantially with changing embedding dimension.

Next, we evaluated the stability of the \textit{direct bias} measure proposed in \autocite{bolukbasiManComputerProgrammer2016}. The authors assert that the direct bias measure can be used as a metric to conclude how much an embedding is biased. For instance, for word embeddings trained on Google News articles, they reported that direct gender bias on 327 profession names is 0.08 and thus they concluded that this embedding is biased. However, as we have illustrated in Figure~\ref{fig:bias_density}, this bias score is not stable, i.e., direct bias measure decays exponentially with increasing word embedding dimension. Moreover, the change in the direct bias, the average bias magnitude in a word embedding, varies more with model hyper-parameter configuration than with corpus selection, and in certain instances, notably Twitter vs. Wikipedia, the bias ranking of corpora may also change with altering embedding dimension. This, in turn, suggests that this bias metric magnitude corresponds less directly to the inherent bias of the text corpus than to to the measurable bias resultant of the embedding training algorithm hyper-parameter selections employed.

\begin{figure}
	\centering
		\includegraphics[width=0.8\textwidth]{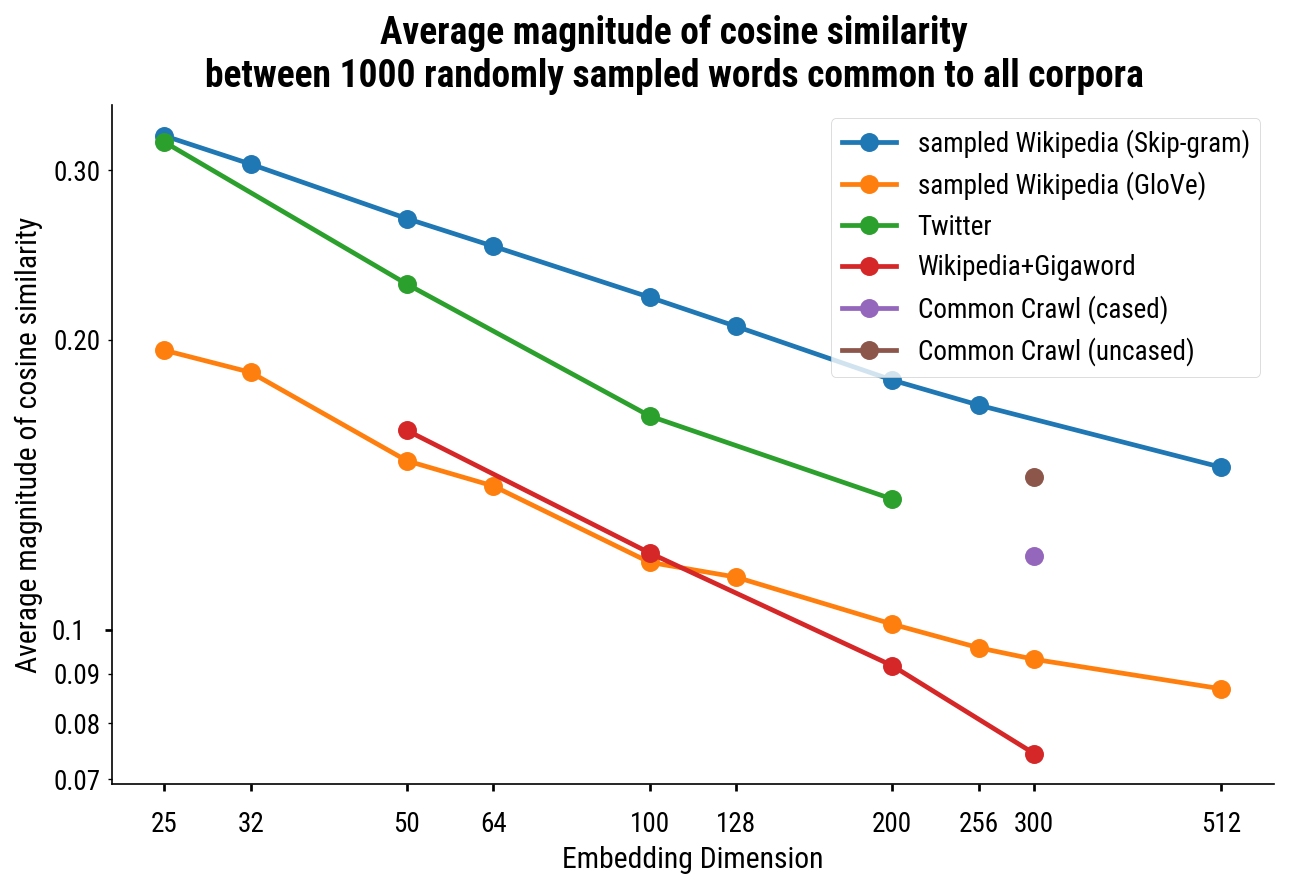}

    \caption{Average magnitude of cosine similarity between 1000 randomly sampled word pairs common to all corpora.}
    \label{fig:average_cosine}
\end{figure}

The decrease of bias measure with increasing embedding dimension could lead to the conclusion that you can reduce direct bias from word embeddings by increasing the dimensionality. However, we believe the downward slope we see in the plots is merely an effect of the properties of the cosine similarity metric in high dimension spaces. In particular, in Figure~\ref{fig:average_cosine} we observe that arbitrary pairs of word embeddings become less and less similar to each other as the number of dimensions increases. We feel additional research regarding whether apparently-low-bias, high dimension embedding spaces regain their measurable bias when projected into smaller dimension spaces wold help further understand this phenomenon. Future work attempting to develop a better canonical bias metric for corpora should seek to be less sensitive to model training hyper-parameters, accounting for the properties of the cosine similarity at various dimensionalities.

\footnotetext[1]{Last 4 embeddings are publicly available at \url{https://nlp.stanford.edu/projects/glove/} }

\begin{figure}
	\centering
		\includegraphics[width=0.8\textwidth]{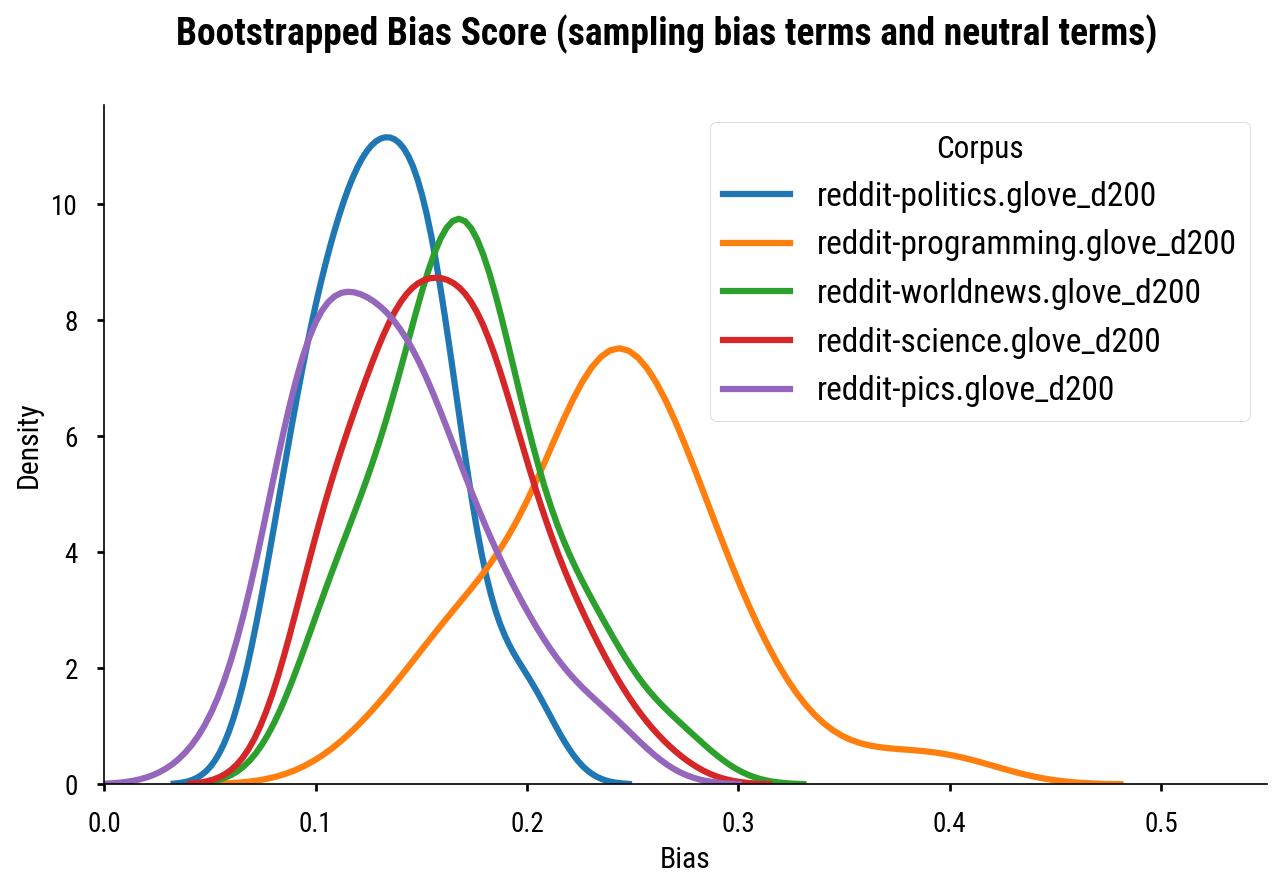}

    \caption{Distributions of direct bias scores of various online corpora under bootstrapping of bias and neutral term sets, $G_1$, $G_2$, and $W$.}
    \label{fig:bootstrap_terms}
\end{figure}

In addition to the explored sensitivity to the dimension of the embedding space, we note that there is some degree of sensitivity of the proposed metric to the sample of terms used to induce the axis onto which term vectors are projected to determine their bias. While it is not difficult to hypothesize various words which are supposed to be either ideally neutral terms $W$, or ideally bias-axis-aligned $G_1$, $G_2$, or as in \autocite{bolukbasiManComputerProgrammer2016}, have these term sets evaluated by a crowd, it is much more difficult to argue the canonicity of a given term set $W$, $G_1$ or $G_2$. If it is not possible to defend the term set selections used to define the metric as being canonical, it is better for them to be mathematically regarded as a sample of the canonical term set. In this light, it becomes critical to understand the sensitivity of the proposed metric to the particular term set sampled, and ideally, characterize the distribution of the metric under many such samples as partially explored in \autocite{antoniakEvaluatingStabilityEmbeddingbased2018}. In examining several online corpora, we observe that although the degree of variance due to term selection at a given embedding dimensionality is small compared to that due to the choice of embedding dimension, we feel that the incorporation of a description of the variance of the bias metric would be prudent to allow for the development of a notion of statistical significance of bias comparisons between corpora. In Figure~\ref{fig:bootstrap_terms}, we show that for several online corpora\footnote[2]{Raw corpora are downloaded from \url{https://files.pushshift.io/reddit/comments/}}, the variance of the bias metric under bootstrap samples of $G_1$, $G_2$, and $W$ is large compared to the inter-group mean differences, making statements comparing these mean bias estimates suspect if reported in absence of an account for the variance of the metric under sampled term sets.

We note that without accounting for the bias metric variance due to target term sample, one would be led to conclude that the bias of \textit{reddit-politics.glove\_d200} is greater than the bias of \textit{reddit-pics.glove\_d200}, while in fact, this difference is probably not significant ($p>0.5$), whereas \textit{reddit-programming.glove\_d200} is probably significantly more biased than \textit{reddit-politics.glove\_d200} ($p<0.001$) even when accounting for the variance of the bias metric due to target term sample.

\section*{Concluding remarks}
\addcontentsline{toc}{section}{Concluding remarks}
We conclude that while meta analyses of the bias metrics proposed by \autocite{bolukbasiManComputerProgrammer2016} indicate that the metrics capture and somewhat quantify sociologically meaningful biases present in learned embedding spaces, the metrics are highly sensitive to the hyper-parameter configurations of the algorithms used to learn them. We specifically find that the average magnitude of the quantified bias is particularly sensitive to the embedding dimension hyper-parameter selected, as well as the sample of bias-axis-inducing terms used to construct the various projections employed by the metric. While it is the case that the bias metrics in \autocite{bolukbasiManComputerProgrammer2016} may provide meaningful rankings of corpora when controlling for model hyper-parameter configuration, publishing the average absolute value of the metric without a complete account for model configuration is suspect. Moreover, we feel publications utilizing these bias metrics would benefit from including information regarding the variance or confidence intervals of the metric under bias-term sampling, if the absolute values of the metric must be published. Regarding the use of the metric as a model selection tool intended to minimize the bias of downstream models employing the embedding space in question, we feel it is important to understand how the properties of the metric employed vary with embedding dimension, and to critically consider whether increased embedding dimensionality truly reduces implicit biases captured within the space, or whether the increased dimensionality simply reduces the apparent magnitude of the bias to simplistic quantification methods per the properties of the cosine similarity in high dimension spaces.

Important steps have been taken regarding the quantification of implicit bias contained in embedding spaces, but we feel there is still effort to be made towards developing metrics which are less sensitive to model hyper-parameter selection, and which possess more robust geometric properties if these metrics are to be used for model selection. When reporting and discussing these metric values for various corpora, we feel it is necessary to include detailed information regarding the embedding learner's hyper-parameter configuration to improve the utility and interpretability of the results.

\printbibliography

\end{document}